\documentclass[fleqn]{VRIH2025}
\usepackage{bm}
\usepackage{multirow}
\usepackage{braket}
\usepackage{algorithm}
\newcommand{\degree}[1]{°\ }
\usepackage{booktabs}
\usepackage{multirow}
\usepackage{xcolor,colortbl}
\newcommand{\gray}{\cellcolor{gray!20}}
\usepackage{acronym}
\usepackage{adjustbox}
\usepackage{orcidlink}
\usepackage{hyperref} 
\newcommand{\hk}[1]{\textcolor{black}{{#1}}}

\begin{document}

\ensubject{Theoretical Physics}
\qyluntan
\ArticleType{Article}
\Year{2025}
\Month{August}
\Vol{68}
\No{8}
\DOI{10.1007/s11433-025-2682-1}
\ReceiveDate{27 January 2025}
\RevisedDate{28 February 2025}
\AcceptDate{28 April 2025}


\title{PanoNormal: Monocular Indoor 360° Surface Normal Estimation}

\author[1]{Kun Huang}{}
\author[2]{Jianwei Yang}{}
\author[2]{Tielin Zhao}{} 
\author[2]{Lei Ji}{}
\author[3]{Songyang Zhang}{{zhangsy@travelsky.com.cn}}
\author[1]{Fang-Lue Zhang}{{fanglue.zhang@vuw.ac.nz}}
\author[1]{Neil Dodgson}{}

\AuthorCitation{K. Huang, J. Yang, T. Zhao, L. Ji, F.L. Zhang, S. Zhang, N.A. Dodgson}

\address[1]{Victoria University of Wellington, Wellington, New Zealand}
\address[2]{CCTEG Mining Technology Research, Beijing, China}
\address[3]{Travelsky Technology Ltd., Beijing, China}

\abstract{The presence of spherical distortion in equirectangular projection (ERP) images presents a persistent challenge in dense regression tasks such as surface normal estimation. Although it may appear straightforward to repurpose architectures developed for 360° depth estimation, our empirical findings indicate that such models yield suboptimal performance when applied to surface normal prediction. This is largely attributed to their architectural bias toward capturing global scene layout, which comes at the expense of the fine-grained local geometric cues that are critical for accurate surface orientation estimation. While convolutional neural networks (CNNs) have been employed to mitigate spherical distortion, their fixed receptive fields limit their ability to capture holistic scene structure. Conversely, vision transformers (ViTs) are capable of modeling long-range dependencies via global self-attention, but often fail to preserve high-frequency local detail. To address these limitations, we propose \textit{PanoNormal}, a monocular surface normal estimation architecture for 360° images that integrates the complementary strengths of CNNs and ViTs. In particular, we design a multi-level global self-attention mechanism that explicitly accounts for the spherical feature distribution, enabling our model to recover both global contextual structure and local geometric details. Experimental results demonstrate that our method not only achieves state-of-the-art performance on several benchmark 360° datasets, but also significantly outperforms adapted depth estimation models on the task of surface normal prediction. The code and model are available at \url{https://github.com/huangkun101230/PanoNormal}.
}


\maketitle

\section{Introduction}\label{sec:introduction}
Estimating surface normal is a crucial task in computer vision as it provides essential geometric information about the structure and orientation of surfaces within a scene. By offering valuable insights into the underlying 3D geometry of objects, accurate surface normal estimation contributes significantly to advancing the capabilities of a wide range of applications, including object recognition, autonomous driving, and robotics. Despite significant progress in normal estimation for conventional perspective images~\cite{long2024adaptive,bae2021estimating,liao2019spherical}, the task of 360\degree\ surface normal estimation remains less explored.  When directly applying the methods proposed for perspective images to 360\degree\ images, the results are unsatisfactory (see Fig.~\ref{fig:d2n}). This is because the spatial warping inherent in 360\degree\ images is not taken into account in these techniques and so we instead, in this paper, develop specialized techniques tailored to their spherical representation.

To address the distortions inherent in panoramic imagery, prior work in related tasks such as 360\degree\ depth estimation has proposed projection–fusion architectures (e.g., UniFuse~\cite{jiang2021unifuse}) that fuse complementary information from equirectangular projection images and cube-map patches to produce more accurate predictions in the spherical domain. While effective for depth, these projection-fusion strategies must reconcile a domain gap between different projections and introduce significant cross-projection fusion overhead. Alternative distortion-aware approaches embed spherical geometry directly into CNNs or adopt specialized loss formulations to compensate for spatial warping~\cite{coors2018spherenet,liao2019spherical}. More recently, self-attention mechanisms have been applied to recover holistic scene structure in the 360\degree\ domain~\cite{li2022omnifusion,shen2022panoformer}, with transformers demonstrating clear advantages in modelling long-range dependencies. However, despite these advances in depth estimation and global modelling, such depth-oriented and transformer-based solutions exhibit two complementary shortcomings when transferred to surface normal estimation: (1) projection-fusion and distortion-aware CNN solutions tend to prioritise correcting large-scale geometric distortions and therefore do not always preserve the high-frequency, local geometric cues required for accurate surface orientation; and (2) conventional transformer pipelines, although powerful at global reasoning, commonly lack inductive mechanisms to retain fine-grained local detail and do not explicitly exploit the spherical distribution of features. These observations — corroborated by our empirical comparisons — motivate a hybrid design that integrates CNN-based local feature extraction with multi-level, spherical-aware self-attention to jointly recover fine-scale geometry and global context for 360\degree\ surface normal estimation.
\begin{figure}[t!]
 \centering
 \includegraphics[width =1\linewidth]{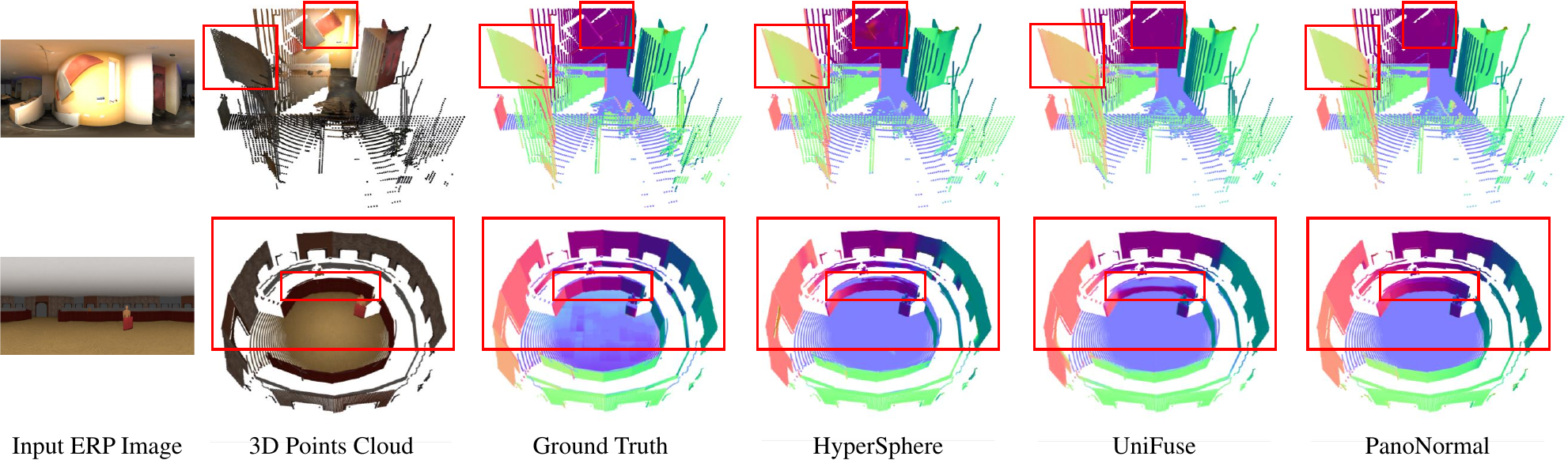}
 \caption{
Our PanoNormal method produces more accurate normal estimation predictions compared to the current state-of-the-art method, particularly in the areas highlighted by the red rectangle. For better visualization, we provide a 3D point cloud generated from the ground truth depth.}
 \label{fig:teaser}
\end{figure}
In this paper, we present PanoNormal, a framework for monocular surface normal estimation in 360\degree\ indoor environments. PanoNormal strategically integrates the strengths of CNNs for extracting low-level features and enhancing locality, along with the advantages of transformers in capturing and associating long-range dependencies. We introduce a deep learning module aimed at extracting low-level features from raw 360\degree\ images, departing from the conventional approach of direct tokenization inputs. To enhance the capabilities of self-attention in exploring the structure of the surface normal map, we propose a multi-scale transformer decoder, enriching the final representation by integrating information across diverse scales. As demonstrated in Fig.~\ref{fig:teaser}, previous state-of-the-art methods yield sub-optimal results when applied to complex scenes, failing to accurately identify complete object boundaries and providing incorrect surface normal vector directions, particularly in the areas highlighted by red boxes. In contrast, our approach predicts sharper and more accurate details of local areas and finely-delineated object boundaries, significantly improving the holistic geometry understanding of the scene. We conduct extensive experiments on public datasets to evaluate our approach, and the results demonstrate that PanoNormal consistently outperforms state-of-the-art models across all examined datasets. Our contributions are summarized as follows:
\begin{itemize}
    \item We present PanoNormal, a specialized vision transformer architecture designed for estimating surface normals in monocular indoor ERP imagery. Notably, PanoNormal stands as the first panoramic transformer tailored for the 360\degree\ surface normal estimation task.
    \item PanoNormal seamlessly integrates the benefits of CNNs for robust low-level feature extraction, fortifying locality. Additionally, it harnesses the strengths of transformers in capturing long-range dependencies. The incorporation of a multi-scale scheme further enhances the holistic representation of the geometric structure within the scene.
    \item Extensive experiments conducted on widely recognized benchmarks (namely 3D60, Stanford2D3D, Matterport3D, SunCG, and Structured3D) illustrate that PanoNormal consistently surpasses state-of-the-art approaches, and contribute to further studies in this domain by providing the first comprehensive evaluation across these public benchmarks.
\end{itemize}

\section{Related Work}\label{sec:relatedwork}
\subsection{360\degree\ Surface Normal Estimation}
Most existing methods~\cite{long2024adaptive,Chen_2023_CVPR,bae2021estimating,do2020surface} for surface normal estimation are designed for perspective images, which poses challenges, such as distortions, deformations, and domain shifts, that make conventional methods unsuitable for 360\degree\ images. Moreover, panoramic images introduce additional complexities to surface normal estimation, including variable illumination and intricate indoor layouts. Recent works can be broadly categorized into direct and indirect methods. Direct methods~\cite{liao2019spherical,karakottas2019360,feng2020deep} introduce specific loss functions (e.g., spherical loss, hyper-sphere loss, double-quaternion loss) to comprehend the geometric structure on the unit sphere with various CNN architectures. Coors et al.~\cite{coors2018spherenet} explicitly embed invariance against spherical distortions into CNNs, adjusting the sampling locations of convolutional filters to counter distortions and align the filters around the sphere effectively. Indirect methods~\cite{albanis2021pano3d,zhao2022monovit} draw inspiration from other vision tasks, e.g., 360\degree\ depth estimation, and adopt their architecture for surface normal estimation. Other work~\cite{li2022omnifusion,jiang2021unifuse,wang2022bifuse++,shen2022panoformer} proposes methodologies that project equirectangular images into corresponding cube map or tangent plane projection sub-images for predictions or directly learn internal correlations among these projections, aiming to capture details and holistic information concurrently with distortion awareness. However, employing simple CNNs with various spherical loss functions, adopting surface normal estimation methods from depth alone, or calculating surface normals directly from depth maps \cite{long2024adaptive} fails to yield satisfactory results. 
Besides, other studies have explored complementary directions in 360° vision. Huang et al.~\cite{huang2023360} and Kou et al.~\cite{kou2024neural,kou2025omniplane} introduced image editing operations specifically designed for 360° representations. Wang et al.~\cite{wang2025target} investigated target scanpaths to enhance 360° media experiences, while Peng et al.~\cite{peng2025robust} utilized dynamic gnomonic projection for object tracking in the 360° domain. Although these works broaden the scope of 360° research, they remain distinct from surface normal estimation. In addition, to the best of our knowledge, there are no surface normal estimation methods that provide a standardized evaluation across diverse public benchmarks, as has been established in depth estimation tasks.

\subsection{Vision Transformer}
Vision transformers are a class of models that use self-attention mechanisms to process images as sequences of patches~\cite{dosovitskiy2020image}. ViTs~\cite{zhao2022monovit,wang2022uformer,jain2023semask,Zhu_2023_CVPR} have achieved remarkable results on conventional images captured by cameras that have a limited field of view and that have, at most, mild distortions. In contrast, 360\degree\ images offer a complete, immersive view of the surrounding scene but require adaptations and modifications of the original ViTs to handle challenges such as significant image distortions, object deformation, and domain shifts. Shen et al.~\cite{shen2022panoformer}, Li et al.~\cite{li2022omnifusion}, Yun et al.~\cite{yun2023egformer}, Zhang et al.~\cite{zhang2025sgformer}, and Ai et al.~\cite{ai2023hrdfuse,ai2024elite360d} proposed extracting features from other projection domains, such as cubemap, tangent, and icosahedron domains, to complement the ERP domain and reduce the negative effects of panoramic distortions, and introduced an attention module that considers both spatial and angular relations among tokens.  Yun et al.~\cite{yun2022panoramic} presented a panoramic ViT for saliency detection in 360\degree\ videos, which leverages a multi-scale feature fusion module and a temporal attention module to capture the spatial and temporal saliency cues. Zhang et al.~\cite{Zhang_2022_CVPR} proposed a method for panoramic semantic segmentation task, which is equipped with deformable patch embedding and deformable MLP modules for handling object deformations and image distortions, and also enhances the mutual prototypical adaptation strategy for unsupervised domain adaptive panoramic segmentation. Recently, multi-task learning for 360° images has gained attention for jointly estimating depth and surface normals. Huang et al. proposed the 360MTL~\cite{huang2025multi}, which employs a spherical-aware Vision Transformer~\cite{shen2022panoformer} to handle distortion and achieved strong performance on both tasks. These works demonstrate the potential and effectiveness of ViTs for 360\degree\ vision tasks and motivated us to investigate suitable ViTs tailored for the surface normal dense prediction task, aiming to seamlessly integrate the strengths of the previously mentioned methods. Our objective is to yield robust evaluation results across benchmarks, contributing to further studies in this domain.

\section{Our Method}\label{sec:method}
\begin{figure*}[t]
 \includegraphics[width =1.0\linewidth]{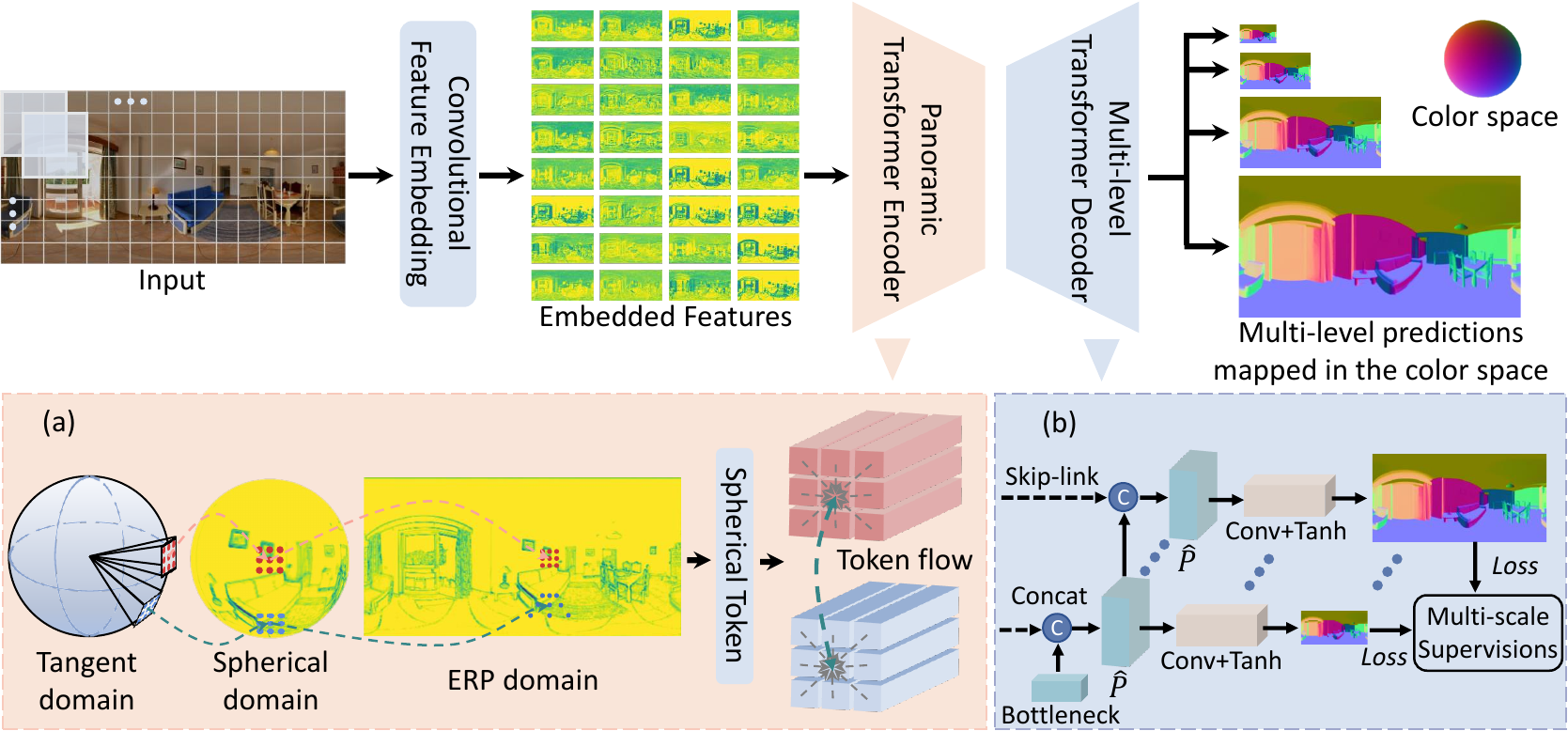}
 \caption{Top: the overall architecture of the proposed PanoNormal method. 
Bottom: the key components: (a) The distortion-aware sampling process on the tangent patch, its transformation to the target ERP domain, and the application of a self-attention scheme among the tokens within each patch. A learnable token flow facilitates attention among the patches. (b) The proposed hierarchical multi-level transformer decoder, which produces results in different scales for comprehensive learning.}
 \label{fig:network}
\end{figure*}
To address the challenge of estimating surface normals using a single panoramic image for indoor scenes, we propose a novel architecture,  PanoNormal. PanoNormal leverages the capabilities of convolutional layers for extracting meaningful embeddings from panoramic images. To address the challenge of spherical distortion in 360\degree\ images while maintaining global dependencies, we crafted a U-shaped distortion-aware transformer architecture, drawing inspiration from the approach proposed by Shen et al.~\cite{shen2022panoformer}. More specifically, we adopt a multi-level structure in our transformer decoder, capturing spatial relationships, processing fine-grained details, and integrating high-level contextual information across different levels. Fig.~\ref{fig:network} illustrates the architecture of PanoNormal.

\subsection{Network Architecture}
\subsubsection{Feature Embedding}
In a conventional ViT-based network, feature embedding undergoes processing through only a single convolutional layer before ViT blocks. It has been demonstrated that this works well for some 360\degree\ understanding tasks, such as depth estimation~\cite{shen2022panoformer}. 
Nonetheless, although both depth estimation and surface normal tasks demand high-quality contextual information, 
it is not effective to apply a single convolutional layer to tackle both depth and surface normal estimation tasks. This is primarily due to the intricate nature of these tasks and their feature requirements. Depth estimation benefits from a broader, more global understanding of the scene, which can be achieved with a single convolutional layer. In contrast, surface normal estimation requires capturing detailed local structures and fine patterns, which is better accomplished with multiple convolutional layers.
To address this issue, we introduce a series of convolutional layers aimed at extracting more effective features. This approach enhances our model's ability to gain a more comprehensive understanding of the entire scene, improving performance in surface normal estimation. The feature embedding block consists of three $3 \times 3$ convolutional layers, each followed by a batch normalization function and a rectifid linear unit (ReLU) activation function.

\subsubsection{Distortion-aware Transformer Encoder}
The inherent non-uniform spatial warping of visual features, caused by spherical distortion in the 360\degree\ image, is built upon the PanoFormer transformer mechanism~\cite{shen2022panoformer}. To address issues such as distortions, misalignment, and inaccuracies arising from direct pixel sampling on the ERP image, a tangent projection division strategy is employed. It samples related surrounding pixels of the central pixel (tangent point) on each dense divided plane in the tangent domain, obtaining a series of tokens in patches with decent position representation that can be further transformed to the ERP domain. Subsequently, a conventional multi-head vision transformer block is applied, with the replacement of the feed-forward network to the locally-enhanced feed-forward network~\cite{yuan2021incorporating,wang2022uformer} for enhancing local feature interaction. 
The introduced encoder not only computes attention scores between central and correlated tokens but also integrates a trainable token flow. This token flow serves as a bias to adjust the spatial distribution of tangent tokens, supplying the network with additional positional information for learning global dependencies.
An example is shown in Fig.~\ref{fig:network}(a) and 
the representation of the self-attention is as follows:
\begin{equation}
    P(f,\hat{s}) = \sum\nolimits_{m}W_m\left[\sum\nolimits_{(q,k)}A_{mqk} \cdot W'_{m}f(\hat{s}_{mqk}+\Delta s_{mqk})\right]
\end{equation}
where the sampling strategy is adopted on the feature representations $f$, and is denoted as $\hat{s}$. The parameters $m, q, k$ represent the self-attention head, each token, and its neighboring tokens in a tangent patch, respectively. $W_m$ and $W'_{m}$ denote the learnable weights of each head, $A_{mqk}$ is the attention weights for each token, and $\Delta s_{mqk}$ signifies the learned bias of each token's query.

The learned embeddings $P$ are subsequently transmitted through skip links to their respective hierarchical decoder blocks. Simultaneously, they undergo downsampling, reducing the feature size by half while doubling the dimensions, before proceeding to a bottleneck block for decoding.

\subsubsection{Multi-level Transformer Decoder} 
Our proposed ViT decoder, designed with spherical distortion awareness and a hierarchical structure to predict normal maps of different scales, is illustrated in Fig.~\ref{fig:network}(b). The proposed multi-level decoder comprises four independent blocks. Each block processes the concatenation of the upsampled encoded representations $\hat{f}_{i}$, featuring twice the spatial resolution size and half the number of channels from the preceding level, and the directly propagated features $\hat{s}_{i}$ from the encoders through the skip-links. The surface normal vectors in various scales are generated as outputs using $3 \times 3$ convolutions, followed by a hyperbolic tangent (tanh) activation function to constrain them to the range $[-1, 1]$. This process can be described as:
\begin{equation}
\hat{\textbf{N}}_i=\tanh\left(\hat{P}(\hat{f}_{i}, \hat{s}_{i})\right)
\end{equation}
where $\hat{\textbf{N}}_i$ denotes the predicted surface normal maps for the $i$-th scale.

 This architecture enhances feature analysis across fine and coarse scales, promoting a holistic comprehension of diverse granularities and elevating spatial understanding. It filters noise effectively, preserving essential features and enhancing generalization to unseen data. The significance of the introduced decoder in enhancing final predicted results has been validated through our ablation study (Sec.~\ref{sec:ablation}). 

\subsection{Loss Function}
The proposed network generates surface normal maps at multiple scales to capture the scene's detailed global geometric structure. During training, the predicted maps are upsampled to match the input size using bilinear interpolation. The training process incorporates MSE, quaternion, perceptual, and smooth loss functions, which are detailed as follows:\\
\textbf{MSE Loss}, $L_{m}$ is the mean squared error between the ground truth and the predicted normals of each pixel, defined as:
\begin{equation}
L_{m} = \sum_{i=1}^{S} \| \hat{\textbf{N}}_i - \textbf{N}_i \|_2
\end{equation}
where $S$ is the number of scales, and $N_i$ denotes the ground truth surface normal maps for the $i$-th scale.\\
\textbf{Quaternion Loss}, $L_{q}$ \cite{karakottas2019360} measures the angular difference between predicted and ground truth normal maps on a pixel-wise basis:
\begin{equation}
    L_{q} = \sum_{i=1}^{N}\sum_{j=1}^{M} \arctan(\frac{\|\hat{\textbf{N}}_{ij} \times \textbf{N}_{ij}\|}{\hat{\textbf{N}}_{ij} \cdot \textbf{N}_{ij}})
\end{equation}
where $M$ indicates the number of pixels of the input, and $j$ indexes the current pixel.
\textbf{Perceptual Loss}, $L_p$ is employed on the finest scales to enhance the generation of finer details:
\begin{equation}
    L_{p}=l_{\textsubscript{feat}}^{\phi,k}(\hat{\textbf{N}},\textbf{N}) = \sum_{j=1}^{M}\frac{1}{C_{k}M}
     \| \phi_{k}(\hat{\textbf{N}}_j) - \phi_{k}(\textbf{N}_j) \|_2^{2}
\end{equation}
where $\phi$ is the VGG-16 network~\cite{simonyan2014very} that pretrained on the ImageNet dataset~\cite{russakovsky2015imagenet}, and $C_{k}$ indicates $C$ dimensional features for the $k$-th layer of the network $\phi$. \\
\textbf{Smooth Loss}, $L_{s}$ quantifies the gradient, $G$, in the $x$ and $y$ directions in the ground truth and the predicted surface normal map at all scales:
\begin{equation}
    L_{s} = \sum_{i=1}^{N}\sum_{j=1}^{M} (| G_{ij}^{x} |+| G_{ij}^{y} |)
\end{equation}
\\
The overall loss function of our network is:
\begin{equation}
    L = \lambda_mL_m + \lambda_qL_q + \lambda_pL_p + \lambda_sL_s
\end{equation}
By default, we set $\lambda _m = 1.0$, $\lambda _q = 10.0$, $\lambda _p = 0.05$, and $\lambda _s = 0.5$ as the weights for different terms. Our experiments show that this specific combination consistently yields the best results compared to other loss configurations. The effectiveness of each loss function is further validated in our ablation study in Sec.  \ref{sec:loss_ablation}

\section{Experiments and Results}\label{sec:results}
We conducted experimental validation using five widely recognized panorama datasets: 3D60~\cite{zioulis2018omnidepth}, Stanford2D3D~\cite{armeni2017joint}, Matterport3D~\cite{chang2017matterport3d}, SunCG~\cite{song2017semantic}, and Structured3D~\cite{zheng2020structured3d}. These datasets were used for both quantitative and qualitative analysis. Additionally, we included the SUN360~\cite{xiao2012recognizing} dataset, which contains real-world data without ground truth, for further qualitative and generalization comparisons.

To evaluate the performance of our method, we compared it against HyperSphere~\cite{karakottas2019360}, which is the current state-of-the-art method specifically designed for 360\degree\ surface normal estimation, and 360MTL~\cite{huang2025multi}, which predicts 360\degree\ surface normals as a sub-task. We also adapted five popular 360\degree\ and perspective image depth estimation methods—PanoFormer~\cite{shen2022panoformer}, OmniFusion~\cite{li2022omnifusion}, MonoViT~\cite{zhao2022monovit}, and UniFuse~\cite{jiang2021unifuse}—by modifying their final prediction layer to output three dimensions instead of one, thereby making them applicable to the surface normal estimation task. This adaptation was crucial for validating the effectiveness of our proposed architecture.

\subsection{Evaluation Metric and Datasets}
We conducted performance evaluation of surface normal estimation using three standard angular error metrics (mean error (Mean), median error (Median), and mean square error (MSE)), and five accuracy metrics that measure the percentage of pixels where the ratio ($\delta$) between the predicted surface normal vector and the ground truth is less than 5°, 7.5°, 11.5°, 22.5°, and 30°. A consistent setting is applied to ensure a fair comparison across all methods, and detailed specifics for each dataset are presented as follows.

\textbf{3D60} dataset offers a broad spectrum of panoramic images with resolutions of $256 \times 512$ captured in varied environments. The captured 360\degree\ RGB imagery with corresponding information, such as surface normal and depth with specific camera positions are from two real-world indoor scanning environments, Stanford2D3D and Matterport3D, alongside synthetic scenes from the SunCG datasets. The inherent distribution gap among these datasets enhances the model's generalizability. To facilitate model training and evaluation, we adopt the data splits utilized in HyperSphere, as recommended in the dataset's documentation. Notably, Matterport3D lacks ground truth data for 360\degree\ surface normals, and Stanford2D3D's surface normal instances lack consistently aligned axes across their data. Consequently, we evaluated them based on specific separations within the 3D60 dataset.

\textbf{Structured3D} is a large-scale synthetic dataset featuring 21,835 RGB images of resolution $512 \times 1024$. These panoramic images were captured across 3500 scenes, illuminated with cold, normal, and warm lighting. The dataset also includes surface normal, depth, and semantic annotations. We preprocessed and formed the dataset examples with an 8:1:1 ratio, yielding 17,442 training data instances with three distinct lighting conditions (52,326 in total with three lighting conditions) and 2,179 and 2,181 validation and test data instances with randomly selected lighting conditions.

\subsection{Implementation Details}
Our experiments were carried out using a single CPU core of an Intel Xeon W-2133 paired with an RTX 3090 GPU, with a batch size configured to 2 and the input resolution set to $256 \times 512$. The chosen optimizer was Adam, with default settings maintained. The initial learning rate was $1\times10^{-4}$, decreasing by half every 15 epochs. We trained our model for 110 epochs and incorporated early stopping at the 15\textsuperscript{th} epoch if there were no further improvements.

\subsection{Experimental Results}
\begin{table*}
    \centering
    \caption{Quantitative comparisons. *Evaluated on corresponding splits of the 3D60 dataset.}

    \resizebox{\linewidth}{!}{
        \begin{tabular}{c|l|rrc|rrrrr}
            \toprule
            \multirow{2}{*}{\centering Dataset} & \multirow{2}{*}{Method}
            & \multicolumn{3}{c|}{Error metric $\downarrow$} & \multicolumn{5}{c}{Accuracy metric $\uparrow$}  \\
            \cline{3-10}
            & & \raisebox{-0.5ex}{Mean}  & \raisebox{-0.5ex}{Median} & \raisebox{-0.5ex}{MSE}  & \raisebox{-0.5ex}{$\delta<$ 5\textdegree} & \raisebox{-0.5ex}{$\delta<$ 7.5\textdegree}  & \raisebox{-0.5ex}{$\delta<$ 11.5\textdegree}  & \raisebox{-0.5ex}{$\delta<$ 22.5\textdegree} & \raisebox{-0.5ex}{$\delta<$ 30\textdegree} \\
            \midrule
            \midrule
    
            \multirow{7}{*}{\centering 3D60}
             & UniFuse&6.7732	&0.5438	&279.9294	&75.67	&78.41	&82.08	&89.27	&92.00\\
             & PanoFormer&17.4997	&6.5841	&929.2326	&50.34	&54.75	&60.28	&72.39	&77.68\\
             & OmniFusion&7.9552	&1.3555	&313.8380	&71.90	&75.57	&79.79	&87.88	&90.96\\
             & MonoViT&6.6087	&0.7666	&250.8495	&75.41	&78.61	&82.57	&89.92	&92.64\\
             &  HyperSphere&5.7836	&\textbf{0.2660}	&224.6234	&76.95	&79.61	&83.49	&90.83	&93.47\\
             &  360MTL &5.4263	&0.3233	&197.8379	&77.75	&80.83	&84.72	&91.53	&94.07\\
            \cline{2-10}
            & \raisebox{-0.4ex}{Ours} &\raisebox{-0.4ex}{\textbf{4.9312}}	&\raisebox{-0.4ex}{0.2671}	&\raisebox{-0.4ex}{\textbf{167.7174}}	&\raisebox{-0.4ex}{\textbf{78.57}}	&\raisebox{-0.4ex}{\textbf{81.70}}	&\raisebox{-0.4ex}{\textbf{85.67}}	&\raisebox{-0.4ex}{\textbf{92.43}}	&\raisebox{-0.4ex}{\textbf{94.89}}\\
            
             & \gray \raisebox{-0.4ex}{Ours-Improve} &\gray\raisebox{-0.4ex}{\textbf{9.12\%}}&\gray\raisebox{-0.4ex}{--0.40\%}	&\gray\raisebox{-0.4ex}{\textbf{15.22\%}}	&\gray\raisebox{-0.4ex}{\textbf{0.82}}	&\gray\raisebox{-0.4ex}{\textbf{0.87}}	&\gray\raisebox{-0.4ex}{\textbf{0.95}}	&\gray\raisebox{-0.4ex}{\textbf{0.90}}	&\gray\raisebox{-0.4ex}{\textbf{0.82}}\\
            \midrule
            \midrule
    
            \multirow{7}{*}{\centering Stanford2D3D*}
             & UniFuse &7.1787	&0.5024	&311.2060	&75.82	&78.27	&81.74	&88.18	&90.96\\
             & PanoFormer &17.5138	&7.3950	&875.6306	&47.81	&52.56	&58.60	&71.68	&77.64\\
             & OmniFusion &8.3031	&1.3415	&336.8175	&71.89	&75.61	&79.81	&87.04	&90.11\\
             & MonoViT &7.1019	&0.7017	&283.0693	&75.27	&78.10	&81.80	&88.59	&91.46\\
             & HyperSphere &6.2644	&\textbf{0.2579}	&257.0234	&76.98	&79.28	&82.80	&89.36	&92.17\\
             &  360MTL&6.0190	&0.3525	&231.6491	&77.21	&79.76	&83.28	&89.90	&92.70\\
            \cline{2-10}
            & \raisebox{-0.4ex}{Ours} &\raisebox{-0.4ex}{\textbf{5.6199}}	&\raisebox{-0.4ex}{0.3213}	&\raisebox{-0.4ex}{\textbf{206.3081}}	&\raisebox{-0.4ex}{\textbf{77.66}}	&\raisebox{-0.4ex}{\textbf{80.26}}	&\raisebox{-0.4ex}{\textbf{83.84}}	&\raisebox{-0.4ex}{\textbf{90.61}}	&\raisebox{-0.4ex}{\textbf{93.44}}\\
             & \gray \raisebox{-0.4ex}{Ours-Improve} &\gray\raisebox{-0.4ex}{\textbf{6.63\%}}	&\gray\raisebox{-0.4ex}{--24.59\%}	&\gray\raisebox{-0.4ex}{\textbf{10.94\%}}	&\gray\raisebox{-0.4ex}{\textbf{0.45}}	&\gray\raisebox{-0.4ex}{\textbf{0.50}}	&\gray\raisebox{-0.4ex}{\textbf{0.56}}	&\gray\raisebox{-0.4ex}{\textbf{0.71}}	&\gray\raisebox{-0.4ex}{\textbf{0.74}}\\
            \midrule
            \midrule
            
            \multirow{7}{*}{\centering Matterport3D*}
             & UniFuse &7.4312	&0.6747	&299.1123	&72.55	&75.69	&79.89	&88.31	&91.38\\
             & PanoFormer &18.3752	&7.4810	&964.8379	&47.26	&51.87	&57.75	&70.82	&76.47\\
             & OmniFusion &8.6668	&1.5917	&337.1754	&68.78	&72.72	&77.41	&86.77	&90.24\\
             & MonoViT &7.2026	&0.9168	&266.8899	&72.36	&76.04	&80.58	&89.08	&92.13\\
             &  HyperSphere &6.3794	&0.3314	&240.0956	&73.79	&76.90	&81.39	&90.04	&93.01\\
             &  360MTL&5.9164	&0.3915	&208.4197	&74.92	&78.56	&83.08	&90.93	&93.76\\
            \cline{2-10}
            & \raisebox{-0.4ex}{Ours} &\raisebox{-0.4ex}{\textbf{5.4004}}	&\raisebox{-0.4ex}{\textbf{0.3182}}	&\raisebox{-0.4ex}{\textbf{177.5569}}	&\raisebox{-0.4ex}{\textbf{75.81}}	&\raisebox{-0.4ex}{\textbf{79.51}}	&\raisebox{-0.4ex}{\textbf{84.13}}	&\raisebox{-0.4ex}{\textbf{91.90}}	&\raisebox{-0.4ex}{\textbf{94.62}}\\
             & \gray \raisebox{-0.4ex}{Ours-Improve} &\gray\raisebox{-0.4ex}{\textbf{8.72\%}}	&\gray\raisebox{-0.4ex}{\textbf{3.99\%}}	&\gray\raisebox{-0.4ex}{\textbf{14.81\%}}	&\gray\raisebox{-0.4ex}{\textbf{0.89}}	&\gray\raisebox{-0.4ex}{\textbf{0.95}}	&\gray\raisebox{-0.4ex}{\textbf{1.05}}	&\gray\raisebox{-0.4ex}{\textbf{0.97}}	&\gray\raisebox{-0.4ex}{\textbf{0.86}}\\
            \midrule
            \midrule
    
            \multirow{7}{*}{\centering SunCG*}
             & UniFuse &3.6620	&0.0430	&170.4886	&88.43	&89.76	&91.47	&94.33	&95.55\\
             & PanoFormer &14.0398	&2.3807	&838.0308	&64.73	&67.96	&71.66	&79.19	&82.50\\
             & OmniFusion &4.6781	&0.3927	&195.2205	&84.84	&87.32	&89.66	&93.27	&94.79\\
             & MonoViT &3.6754	&0.2126	&153.2902	&88.11	&89.70	&91.55	&94.65	&95.88\\
             &  HyperSphere &2.8576	&0.0035	&129.4394	&89.99	&91.16	&92.81	&95.51	&96.59\\
             &  360MTL&2.8032	&0.0072	&119.2599	&89.99	&91.27	&92.93	&95.64	&96.73\\
            \cline{2-10}
            & \raisebox{-0.4ex}{Ours}&\raisebox{-0.4ex}{\textbf{2.3221}}	&\raisebox{-0.4ex}{\textbf{0.0027}}	&\raisebox{-0.4ex}{\textbf{89.3309}}	&\raisebox{-0.4ex}{\textbf{90.86}}	&\raisebox{-0.4ex}{\textbf{92.13}}	&\raisebox{-0.4ex}{\textbf{93.78}}	&\raisebox{-0.4ex}{\textbf{96.41}}	&\raisebox{-0.4ex}{\textbf{97.43}}\\
             & \gray \raisebox{-0.4ex}{Ours-Improve} &\gray\raisebox{-0.4ex}{\textbf{17.16\%}}	&\gray\raisebox{-0.4ex}{\textbf{23.62\%}}	&\gray\raisebox{-0.4ex}{\textbf{25.10\%}}	&\gray\raisebox{-0.4ex}{\textbf{0.87}}	&\gray\raisebox{-0.4ex}{\textbf{0.86}}	&\gray\raisebox{-0.4ex}{\textbf{0.85}}	&\gray\raisebox{-0.4ex}{\textbf{0.77}}	&\gray\raisebox{-0.4ex}{\textbf{0.70}}\\
            \midrule
            \midrule
            
            \multirow{7}{*}{\centering Structured3D}
             & UniFuse &8.2525	&0.3342	&453.1296	&76.24	&81.43	&83.52	&87.55	&89.54\\
             & PanoFormer &16.9159	&4.4362	&1053.5312	&59.13	&64.10	&68.29	&75.50	&78.86\\
             & OmniFusion &20.7000	&14.3028	&831.9758	&28.51	&35.35	&45.06	&63.55	&72.52\\
             & MonoViT &5.9222	&\textbf{0.1005}	&277.8236	&78.93	&84.20	&86.40	&90.58	&92.57\\
             &  HyperSphere &5.7865	&0.1410	&253.3796	&78.38	&83.68	&86.12	&90.73	&92.88\\
             &  360MTL&9.0599	&0.5036	&476.3303	&72.42	&77.78	&80.56	&85.92	&88.49\\
            \cline{2-10}
            & \raisebox{-0.4ex}{Ours} &\raisebox{-0.4ex}{\textbf{5.5622}}	&\raisebox{-0.4ex}{0.1048}	&\raisebox{-0.4ex}{\textbf{246.5729}}	&\raisebox{-0.4ex}{\textbf{79.18}}	&\raisebox{-0.4ex}{\textbf{84.48}}	&\raisebox{-0.4ex}{\textbf{86.68}}	&\raisebox{-0.4ex}{\textbf{91.01}}	&\raisebox{-0.4ex}{\textbf{93.08}} \\
             & \gray\raisebox{-0.4ex}{Ours-Improve}  &\gray\raisebox{-0.4ex}{\textbf{3.88\%}}	&\gray\raisebox{-0.4ex}{--4.28\%}	&\gray\raisebox{-0.4ex}{\textbf{2.69\%}}
    &\gray\raisebox{-0.4ex}{\textbf{0.79}}	&\gray\raisebox{-0.4ex}{\textbf{0.80}}	&\gray\raisebox{-0.4ex}{\textbf{0.55}}	&\gray\raisebox{-0.4ex}{\textbf{0.28}}	&\gray\raisebox{-0.4ex}{\textbf{0.19}} \\
            \midrule
            \midrule
            \multicolumn{2}{c|}{Ours Average Improvement}&{\textbf{6.50\%}}	&{--2.34\%}	&{\textbf{8.96\%}}	&{\textbf{0.81}}	&{\textbf{0.84}}	&{\textbf{0.75}}	&{\textbf{0.59}}	&{\textbf{0.51}} \\
            \bottomrule
        \end{tabular}
    }    
    \label{tab:quantitative_comparison}
\end{table*}

We present a comprehensive quantitative comparison between state-of-the-art methods (HyperSphere and 360MTL), algorithms adapted for spherical surface normal estimation, and our PanoNormal model across five datasets, as detailed in Table~\ref{tab:quantitative_comparison}. For a fair evaluation, all models were retrained under identical settings. PanoNormal demonstrates superior performance, achieving state-of-the-art results across all five benchmarks. On average, it delivers a consistent improvement of 6.50\% in mean error and 8.96\% in MSE, surpassing the previous best methods highlighted in grey.

In specific cases, our model shows significant improvements in the MSE metric: 15.22\% on 3D60, 10.94\% on Stanford2D3D, 14.81\% on Matterport3D, and 25.10\% on SunCG. However, the improvement is relatively modest on the Structured3D dataset, with only a 2.69\% enhancement in MSE. This less significant difference is due to the complex, synthetic nature of Structured3D scenes, which contain numerous small objects with subtle curvature changes, such as ornaments, drawer handles, and wall-mounted kitchen tools. Additionally, the dataset challenges the models with detailed textures in items like mirrors, glass walls, and carpets. 

\begin{figure*}[t]
 \centering
 \includegraphics[width =1\linewidth]{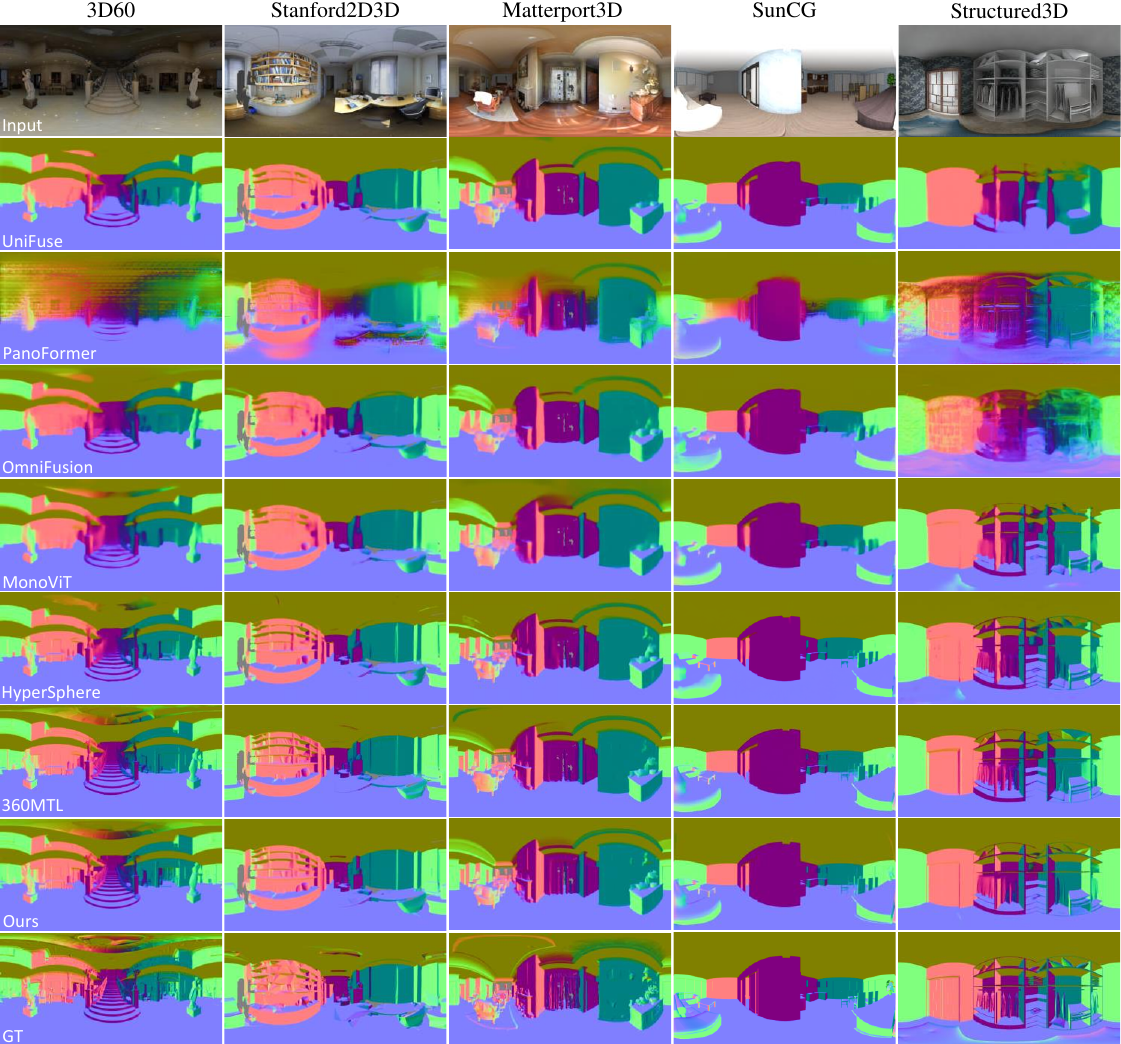}
 \caption{\textbf{Qualitative comparisons across five benchmarks}, featuring PanoNormal, UniFuse, PanoFormer, OmniFusion, MonoViT, HyperSphere, and 360MTL. Optimal viewing experience in color.}
 \label{fig:quali}
\end{figure*}

In addition, PanoNormal shows a marginally higher median error than HyperSphere on the Stanford2D3D dataset, indicating potential sensitivity to outliers. However, this discrepancy also suggests that our method captures finer details. HyperSphere, on the other hand, sacrifices sharp boundaries in its predictions, resulting in more blurred outcomes, as illustrated in Fig.~\ref{fig:quali}.The improvement in $\delta$ performance underscores PanoNormal's ability to achieve more accurate predictions, demonstrating its generalizability and effectiveness across diverse domains, including both real-world and synthetic scenarios within the datasets.

We also provide qualitative results, showcasing RGB input images alongside the corresponding predictions generated by various methods, with the ground truth surface normal map shown in Fig.~\ref{fig:quali}. Each dataset's test example is displayed with specific colors mapped to coordinate information, offering an intuitive representation of surface normals. Areas with unavailable data are denoted by grey on the surface normal map. The PanoNormal model stands out for capturing finer details and sharper boundaries, showing greater sensitivity to subtle surface curvatures.

\begin{figure}[t!]
 \centering
 \includegraphics[width =1\linewidth]{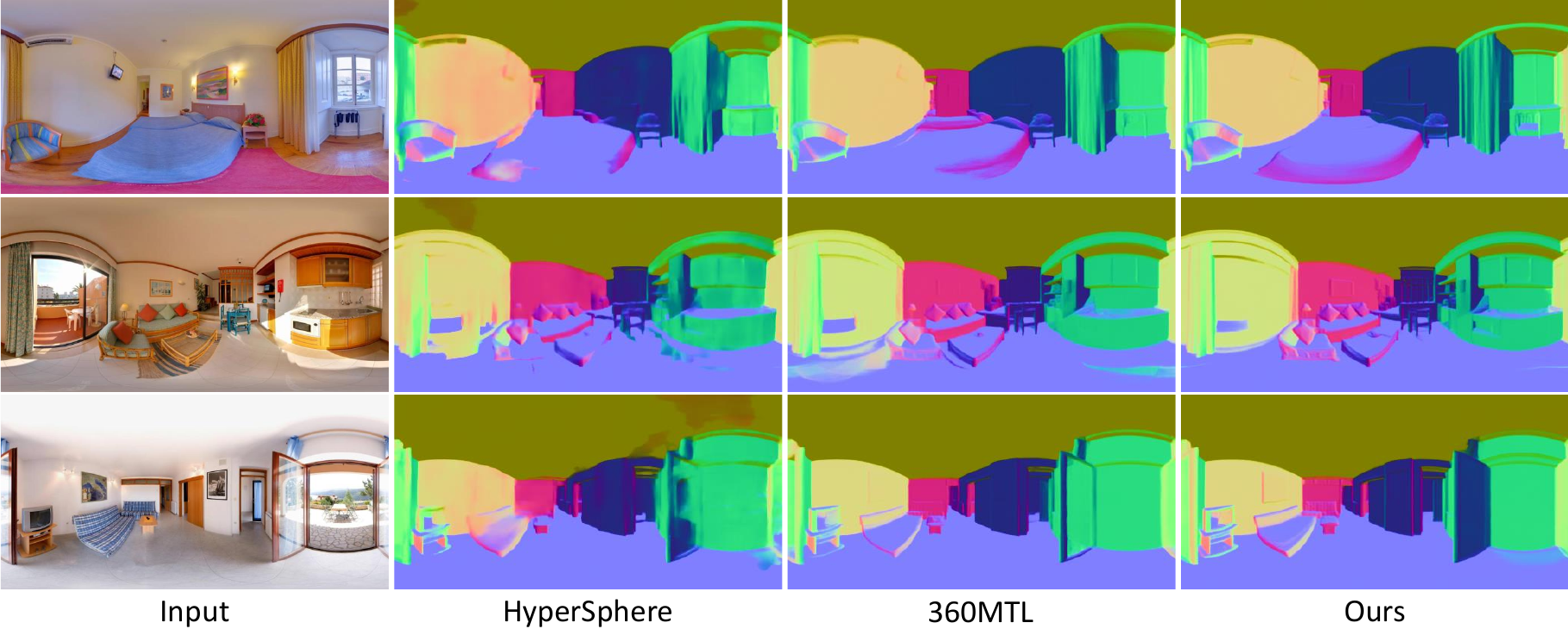}
 \caption{Normal estimation predictions on some real-world data. The images are from the SUN360~\cite{xiao2012recognizing} dataset. More qualitative results can be found in our supplementary materials.}
 \label{fig:realworld_fig2}
\end{figure}

To assess the generalizability of our method, we conduct surface normal estimation on real-world data, comparing our results with those from HyperSphere and 360MTL. As demonstrated in Fig.~\ref{fig:realworld_fig2}, PanoNormal produces precise predictions on the unseen real-world SUN360 dataset, accurately estimating the complete boundaries of various objects even in the presence of distortions and complex textures. Noteworthy examples include the bed at the bottom of the ERP image in the first row and the glass door in the third row. In contrast, HyperSphere does not generalize well to real-world panoramic images, further validating the efficacy of our proposed method.

When examining Stanford2D3D and Matterport3D as real-world datasets, it is important to note that their ground truth surface normal maps are constructed with triangle faces, which may introduce certain visual characteristics. However, our method shows a strong ability to generate smooth surface normals with enhanced boundary precision, contributing to a higher standard of visual quality in our results. More discussion are available in the supplementary material.

\begin{figure}[t!]
 \centering
 \includegraphics[width =1.0\linewidth]{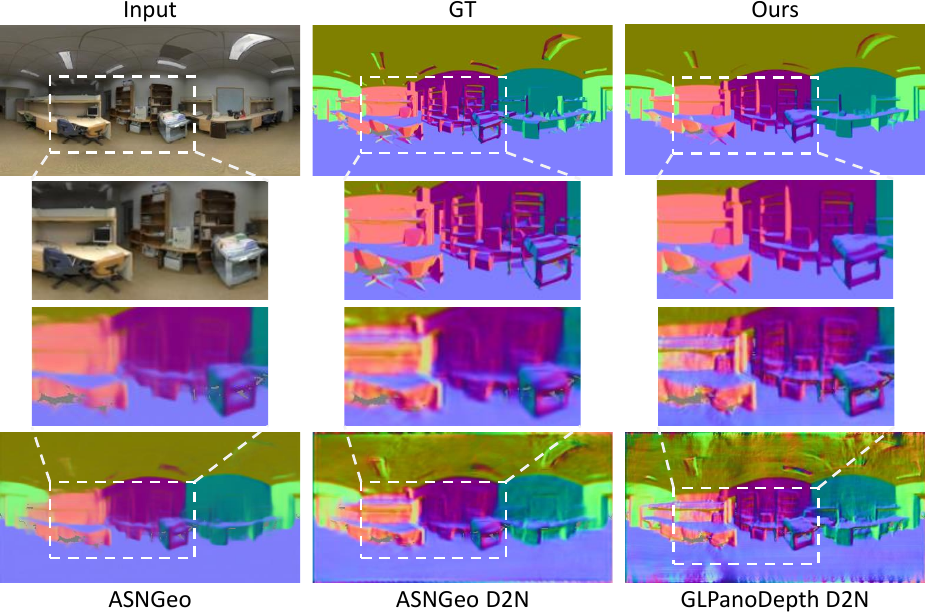}
 \caption{
 Surface normal estimation predictions on Stanford2D3D data are compared between ASNGeo~\cite{long2024adaptive} (a perspective-based method) and our approach. To further validate the conversion process, we used the depth-to-surface-normal technique from ASNGeo~\cite{long2024adaptive} to convert both perspective depth maps and GLPanoDepth \cite{bai2024glpanodepth} (360\degree\ domain) depth maps into surface normals.
 }
 \label{fig:d2n}
\end{figure}

To evaluate the effectiveness of applying perspective image algorithms to the 360° domain, we conduct experiments using ASNGeo~\cite{long2024adaptive}, a recent multi-task model designed to predict both depth and surface normal maps for perspective images. ASNGeo includes a depth-to-normal (D2N) layer that calculates surface normals by averaging directions from multiple randomly sampled triangle faces covering each 3D point. We retrained ASNGeo using 360° data, keeping the default settings for the D2N layer to maintain consistency with the original paper. As shown in Fig. \ref{fig:d2n}, the qualitative results from ASNGeo’s surface normal prediction branch reveal suboptimal performance, even when objects in the ERP image exhibit minimal distortion. We also apply the D2N layer to depth maps from both ASNGeo and a recent 360° depth estimation approach (GLPanoDepth \cite{bai2024glpanodepth}). The results show blurry outputs around objects and inconsistent appearances for ceilings and floors. Additionally, deriving surface normals from depth using this method accumulates errors, ultimately leading to poorer results. These highlight the necessity of developing specialized algorithms for 360° surface normal estimation.

\subsection{Ablation Study}\label{sec:ablation}
\begin{table}[t!]
    \centering
    \caption{Ablation study for individual components. Beginning with the PanoFormer baseline, which has no feature embedding and a multi-level decoder, we incrementally add each component one at a time.}    
    \begin{adjustbox}{width =0.7\linewidth}

	\begin{tabular}  {l|ccc}
			\toprule 
			Method & Mean~$\downarrow$  &Median~$\downarrow$ &MSE~$\downarrow$\\
			\midrule
			 Baseline (\textbf{B}) & 17.500 & 6.584 & 929.233 \\
			 B + multi-level decoder &5.154 & 0.321 & 178.146 \\
			 B + FeatEmbed + multi-level decoder  & 4.931 & 0.267 & 167.717 \\
			\bottomrule 
	\end{tabular}
    \end{adjustbox}

    \label{tab:ablation_component}
\end{table}

\subsubsection{Individual component study.}
We conducted an ablation study to validate the key components of our model. As presented in Tab.~\ref{tab:ablation_component}, PanoFormer serves as the baseline structure for surface normal prediction. Originally designed for depth estimation, PanoFormer exhibits suboptimal performance when directly applied to surface normals, primarily due to the distinct nature of the tasks. Surface normal estimation relies on fine-grained local geometric details, whereas depth estimation emphasizes global scene structure, highlighting the architectural mismatch when the baseline is used without modification.

To address these limitations we introduce, first, a multi-level decoder, and then add a feature embedding module. The multi-level decoder was particularly impactful, resulting in significant reductions in mean, median, and MSE error metrics by \textbf{70.55\%}, \textbf{95.12\%}, and \textbf{80.83\%}, respectively. We believe that its effectiveness arises from its ability to aggregate multi-scale information, combining low-level features for detailed geometry with high-level contextual cues, which is crucial for accurately predicting surface normals across diverse regions, including areas with sharp boundaries or fine textures. This capability ensures that both global consistency and local precision are maintained, addressing one of the primary shortcomings of the baseline.

Building on this, the feature embedding module was designed to enhance the representation of fine local geometric structures, capturing intricate surface characteristics such as subtle curvature changes and texture discontinuities that are often overlooked in global feature representations. Integrating this module into the architecture led to further improvements, with reductions of 4.32\% in the mean error, \textbf{16.84\% in the median error}, and 5.85\% in MSE. The module effectively complements the multi-level decoder by focusing on local feature refinement, enabling the model to perceive nuanced geometric variations in complex scenes.

Together, these components address three critical factors: the task-specific mismatch in the baseline architecture, the need for effective multi-scale feature aggregation, and the importance of capturing detailed local geometry. By leveraging these enhancements, our model achieves a significant improvement in surface normal prediction accuracy, demonstrating the value of balancing global context with fine-grained detail. These findings underscore the necessity of architectural adaptations for domain-specific tasks like surface normal estimation and the potential of our approach to set a new benchmark in this field.

\begin{table}[t!]
    \centering
    \caption{Ablation study for loss combinations. Starting with the MSE and perceptual loss, we progressively incorporate the quaternion and smooth loss.}
    \begin{adjustbox}{width =0.6\linewidth}
		\begin{tabular}  {l|ccc}
			\toprule 
			Loss & Mean~$\downarrow$  &Median~$\downarrow$ &MSE~$\downarrow$\\
			\midrule
			 $L_{m}$ + $L_{p}$ & 5.277 & 0.593 & 170.768 \\
			 $L_{m}$ + $L_{p}$ + $L_{q}$ &5.130 & 0.314 & 179.481 \\
			 $L_{m}$ + $L_{p}$ + $L_{s}$ & 5.421 & 0.660 & 174.738 \\
			 $L_{m}$ + $L_{p}$ + $L_{q}$ + $L_{s}$ & 4.931 & 0.267 & 167.717 \\
			\bottomrule 
	\end{tabular}
    \end{adjustbox}
    \label{tab:ablation_loss}
\end{table}

\begin{table}[t!]
    \centering
        \caption{\hk{Ablation study evaluating the effect of varying the number of ViT and convolutional blocks in our model. We also report comparison with HyperSphere.}}
        \begin{adjustbox}{width =1\linewidth}
		\begin{tabular}  {lcc|ccc|ccc}
			\toprule 
			Method & $\#$ViT block & $\#$Conv block &GFLOPs~$\downarrow$ &Params (M)~$\downarrow$ &FPS~$\uparrow$ & Mean~$\downarrow$  &Median~$\downarrow$ &MSE~$\downarrow$\\
			\midrule
			 Ours & 9 & 0 & 179.60 & 35.14 &10.13 & 4.931 & 0.267 & 167.717 \\
			 Ours-V1 & 7 & 2 & 172.04 & 35.09 & 20.65 & 5.310 & 0.394 & 183.356 \\
			 Ours-V2 & 5 & 4 & 185.50 & 34.98 & 32.19 & 5.337 & 0.383 & 186.472 \\
			 Ours-V3 & 3 & 6 & 211.80 & 34.72 & 41.30 & 5.305 & 0.358 & 184.613 \\
			 Ours-V4 & 1 & 8 & 223.68 & 30.16 & 53.39 & 5.506 & 0.398 & 193.994 \\
			 Ours-V5 & 0 & 9 & 224.50 & 30.98 & 60.81 & 5.533 & 0.360 & 202.060 \\
			\midrule
              HyperSphere & - & - & 197.94 & 35.12 & 68.41 & 5.784 & 0.266 & 224.623\\
			\bottomrule 
	\end{tabular}
    \end{adjustbox}
    \label{tab:ablation_PFblock}
\end{table}

\subsubsection{Loss combinations.}
\label{sec:loss_ablation}
We investigate the effectiveness of various losses on our model's accuracy to determine the optimal combination for training. As shown in Tab.~\ref{tab:ablation_loss}, we began with the common losses: MSE ($L_{m}$) and perceptual loss ($L_{p}$), on the 3D60 dataset. Adding the quaternion loss ($L_{q}$) improved the mean and median error metrics by 2.79\% and 47.05\%, respectively, though it increased the MSE error, indicating higher sensitivity to outliers. Conversely, adding the smooth loss ($L_{s}$) alone decreased performance. However, combining all four losses with specific weights yielded the best results. This suggests that while the quaternion loss sharpens scene details but is sensitive to outliers, incorporating smooth loss helps achieve a balance, enhancing fine-grained details without sensitivity to outliers.

\subsubsection{Number of ViT blocks}
The number of ViT blocks affects both the accuracy and efficiency of our method. \hk{To investigate this, we progressively replaced pairs of self-attention blocks with convolutional blocks in both the encoder and decoder, resulting in multiple configurations with different ViT-to-convolution ratios (Tab.~\ref{tab:ablation_PFblock}). The results show that accuracy consistently decreases as more convolutional layers are substituted, confirming the effectiveness of ViT blocks for capturing long-range dependencies. The best performance is obtained when all blocks are ViTs. Interestingly, the variant ``Ours-V3,'' which retains self-attention only at the bottleneck and one higher level in both encoder and decoder, achieves competitive accuracy while significantly simplifying the architecture. This indicates that a small number of strategically placed ViT blocks can still preserve most of the benefits of global reasoning, while potentially making the model more suitable for deployment on resource-constrained devices.}

\subsubsection{Efficiency of our models}
\hk{In addition to accuracy, we report computational cost (GFLOPs), model size, and inference speed (frame per second, FPS) in Tab.~\ref{tab:ablation_PFblock}. These results highlight a clear trade-off between accuracy and efficiency: ViT-heavy configurations deliver the best error metrics, whereas conv-heavy variants run faster and require fewer resources. This flexibility allows our framework to be adapted to different application requirements, such as prioritizing accuracy in offline high-fidelity settings or emphasizing speed in real-time scenarios. Compared to HyperSphere, our models offer a more favorable balance, achieving comparable or better accuracy with competitive computational efficiency. For instance, mid-range variants like Ours-V3 provide a good compromise, maintaining strong predictive performance while reducing memory usage and improving inference speed, making them attractive candidates for practical deployment.}

\section{Limitation}
\label{sec:limitation}
In Tab.~\ref{tab:quantitative_comparison}, our method exhibits a performance of approximately 24.59\% higher median error compared to HyperSphere on the Stanford2D3D dataset while demonstrating significant improvements in the metrics of mean and MSE. The enhancements on mean and MSE suggest that, on average, our model provides more accurate predictions. However, the higher median error implies the existence of a subset of predictions with larger errors.

\begin{table}[!]
    \centering
        \caption{Specific examples from Stanford2D3D dataset.}
        \begin{adjustbox}{width =0.6\linewidth}
        \begin{tabular}  {c|c|rcc}
			\toprule 
			Example & Method & Mean~$\downarrow$  &Median~$\downarrow$ &MSE~$\downarrow$ \\
			\midrule
              \multirow{2}{*}{\centering (a)}&
			 HyperSphere & 10.5388 & 0.0656 & 470.1378 \\  
			 & Ours & 10.4650 & 0.0816 & 450.3420 \\
		   \midrule
              \multirow{2}{*}{\centering (b)}&
			 HyperSphere & 16.8774 & 7.6914 & 753.1895 \\  
			 & Ours & 16.1840 & 8.8487 & 634.5417 \\
		   \midrule
              \multirow{2}{*}{\centering (c)}&
			 HyperSphere & 13.2489 & 0.8248 & 564.9490 \\   
			 & Ours & 12.2771 & 2.3333 & 473.4669 \\
			\bottomrule 
	\end{tabular}
 \end{adjustbox}
    \label{tab:limitation}
\end{table}

\begin{figure*}[t]
    \centering
 \includegraphics[width =1.0\linewidth]{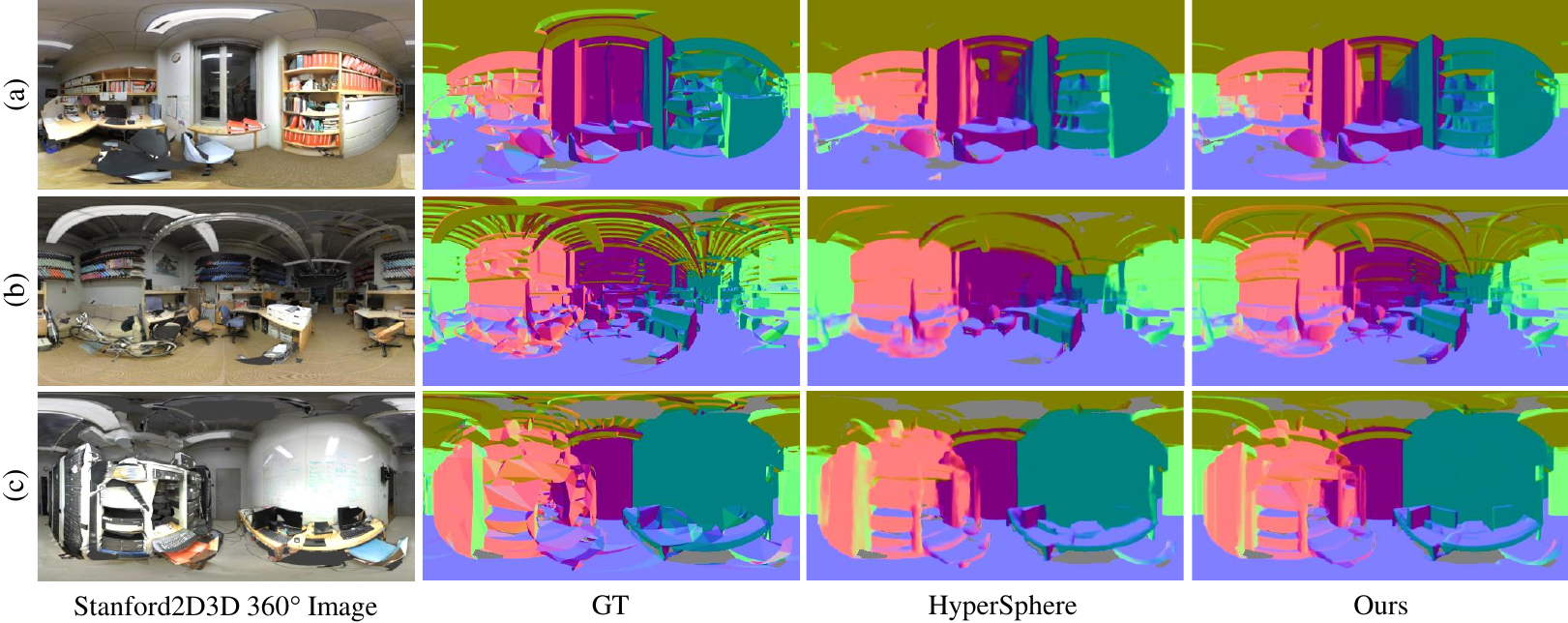}
 \caption{Qualitative evaluation on three specific examples. Note that the Ground Truth is based on triangular faces, which is not realistic. HyperSphere achieves lower median error but blurs object boundaries, while our method predicts more precise boundaries that align better with the actual shape of objects.}
 \label{fig:limitation}
\end{figure*}

To investigate the factors contributing to the higher median error, we inspected the predicted results. We present three typical cases from the Stanford2D3D dataset, both quantitatively in Tab.~\ref{tab:limitation} and qualitatively in Fig.~\ref{fig:limitation}. Notably, the ground truth surface normal data in the Stanford2D3D dataset is generated based on the reconstructed triangular mesh of each real-world scene.  Predictions aligning with surfaces constructed with triangles may deviate from the authentic appearance of objects. HyperSphere, in such instances, blurs object boundaries and achieves a lower median error on such regions, while our approach predicts a more precise boundary that better aligns with the actual shape of objects. This distinction contributes to the overall higher median error, attributed to the inherent limitations in the quality of the ground truth data.

\section{More Comparison on SUN360}

\begin{figure*}[t]
    \centering
 \includegraphics[width =1.0\linewidth]{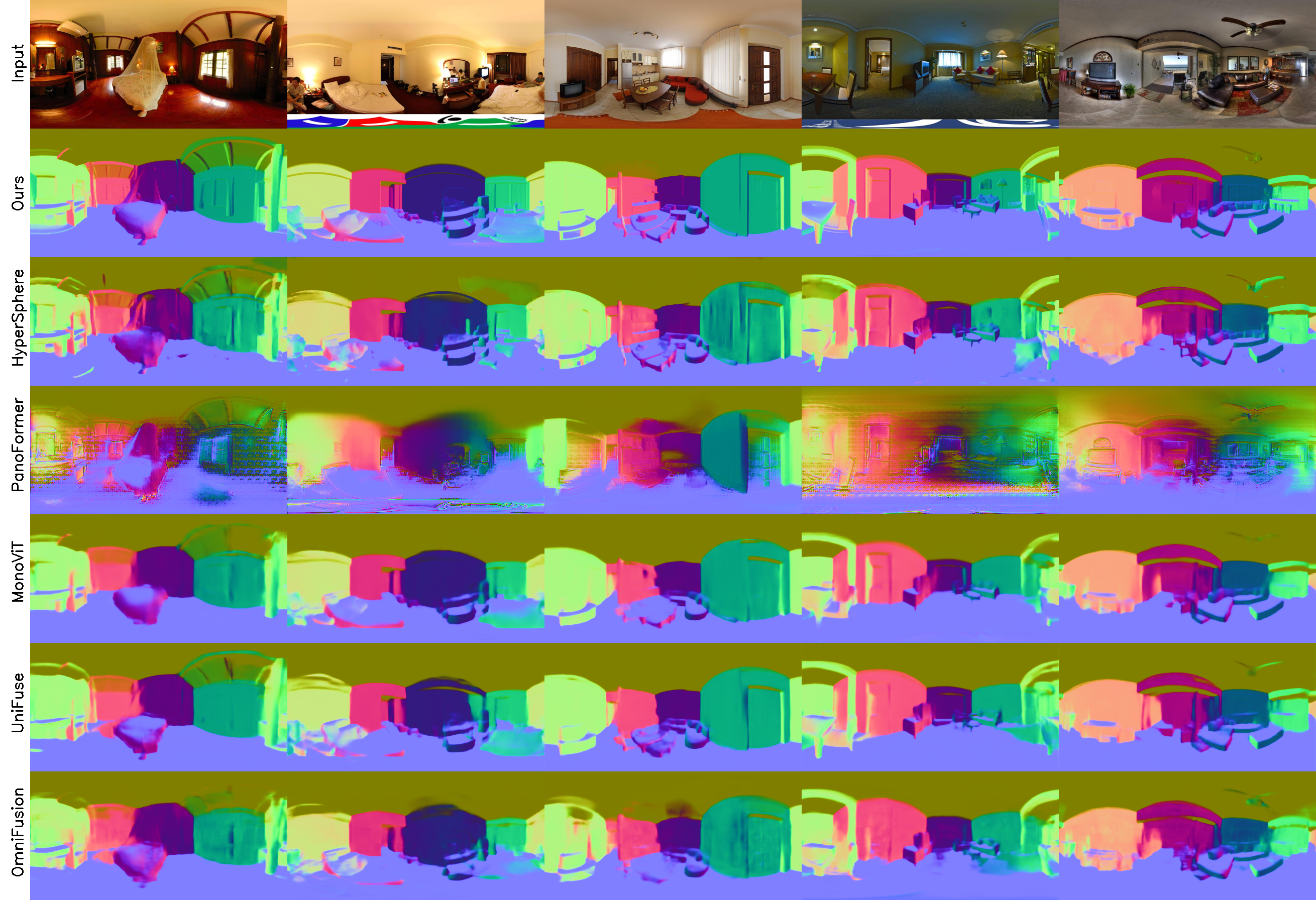}
 \caption{
 Qualitative comparison of the models for the SUN360 dataset. Please note that the SUN360 dataset is a real-world dataset and does not include ground truth data. We provide qualitative comparison of the models across the entire SUN360 which can be found in the separated supplementary materials.
 }
 \label{fig:sun360}
\end{figure*}

In Fig.~\ref{fig:sun360}, we show qualitative comparisons on the SUN360 dataset for 360 surface normal estimation. Our model demonstrates superior performance compared to existing state-of-the-art approaches, includes models such as HyperPSphere, PanoFormer, MonoViT, UniFuse, and OmniFusion, all of which have been evaluated on diverse indoor scenes.

Our model consistently delivers more accurate and coherent surface normal predictions, as evident from the sharpness and continuity of the estimated surfaces across varying room structures. Unlike competing models, which often produce fragmented or blurred normal maps, especially in areas with complex geometries, our approach maintains clarity and precise border of surfaces. This is particularly noticeable in regions with intricate details, such as furniture edges and room corners, where other methods tend to falter.

These results highlight the robustness and effectiveness of our method in handling the challenges posed by real-world 360° imagery. Our method excels in maintaining visual consistency and preserving fine details across various indoor environments, outperforming other models in the process. The ability of our model to produce clear and accurate surface normal estimations, even in complex scenes, sets a new standard for 360 surface normal estimation and establishes it as a reliable solution for applications that demand a high level of environmental understanding.

\section{Conclusion}\label{sec:conclusion}
We present PanoNormal, a deep architecture designed for single 360\degree\ surface normal estimation. Our approach leverages the combined strengths of convolutional neural networks for effectively capturing local details and spherical vision transformers for global dependency with distortion awareness. Experiments demonstrate PanoNormal's significant advances over state-of-the-art methods across five popular 360\degree\ panoramic datasets. Additionally, PanoNormal demonstrates enhanced generalizability compared to previous works, with our ablation studies further validating the effectiveness of our proposed architecture. Our findings also reveal that models originally designed for 360\degree\ depth estimation are not directly transferable to surface normal prediction, underscoring the necessity of tailored architectural design. This makes it a reliable tool for tasks that demand precise environmental understanding.

\section*{Acknowledgements}

This research was supported by the Marsden Fund Council managed by the Royal Society of New Zealand (No. MFP-20-VUW-180) and Tiandi Science and Technology Co. through the Technology Collaboration Project (No. 2025-TD-CXY003).

\InterestConflict{The authors declare that they have no conflict of interest.}

\AuthorContributions{\textbf{Kun Huang:} Conceptualization, Paper writing, Method design, Experiment design \& conduct. \textbf{Jianwei Yang:} Experiment design, Data processing. \textbf{Tielin Zhao:} Experiment design, Data processing. \textbf{Lei Ji:} Experiment design, Data processing. \textbf{Songyang Zhang:} Experiment design, Data processing. \textbf{Fang-Lue Zhang:} Method design, Experiment design and Paper writing. \textbf{Neil Dodgson:} Method design, Experiment design and Paper writing.}

\bibliographystyle{splncs04}
\bibliography{main}

\end{document}